\documentclass[10pt,twocolumn,letterpaper]{article}

\usepackage{iccv}
\usepackage{times}
\usepackage{epsfig}
\usepackage{graphicx}
\usepackage{amsmath}
\usepackage{amssymb}

\usepackage{bm,bbm}
\usepackage{algorithm}
\usepackage{algorithmic}
\usepackage{graphicx}
\usepackage{tabularx}
\usepackage{array,arydshln}
\usepackage{subcaption}

\usepackage{color}

\makeatletter
\newcommand*{\rom}[1]{\expandafter\@slowromancap\romannumeral #1@}
\makeatother

\usepackage[pagebackref=true,breaklinks=true,letterpaper=true,colorlinks,bookmarks=false]{hyperref}

\iccvfinalcopy 

\def\iccvPaperID{924} 

\ificcvfinal\fi
\begin{document}

\title{Learning like a Child: \\
Fast Novel Visual Concept Learning from Sentence Descriptions of Images}

\author{
Junhua Mao$^1$\ \ \ \ \ Wei Xu$^2$ \ \ \ \ \ Yi Yang$^2$ \ \ \ \ \ Jiang Wang$^2$ \ \ \ \ \ Zhiheng Huang$^2$ \ \ \ \ \ Alan L. Yuille$^1$\\
$^1$University of California, Los Angeles \ \ \ \ \ \ \ $^2$Baidu Research\\
{\fontsize{9}{10}\selectfont \tt{mjhustc@ucla.edu}, \tt{\{wei.xu,yangyi05,wangjiang03,huangzhiheng\}@baidu.com},  \tt{yuille@stat.ucla.edu}}
}

\maketitle

\begin{abstract}
In this paper, we address the task of learning novel visual concepts, and their interactions with other concepts, from a few images with sentence descriptions.
Using linguistic context and visual features, our method is able to efficiently hypothesize the semantic meaning of new words and add them to its word dictionary so that they can be used to describe images which contain these novel concepts.
Our method has an image captioning module based on \cite{mao2014deep} with several improvements.
In particular, we propose a transposed weight sharing scheme, which not only improves performance on image captioning, but also makes the model more suitable for the novel concept learning task.
We propose methods to prevent overfitting the new concepts. 
In addition, three novel concept datasets are constructed for this new task, and are publicly available on the project page.
In the experiments, we show that our method effectively learns novel visual concepts from a few examples without disturbing the previously learned concepts.
The project page is: \url{www.stat.ucla.edu/~junhua.mao/projects/child_learning.html}.
\end{abstract}
\vspace{-0.5cm}

\section{Introduction}
Recognizing, learning and using novel concepts is one of the most important cognitive functions of humans.
When we were very young, we learned new concepts by observing the visual world and listening to the sentence descriptions of our parents. The process was slow at the beginning, but got much faster after we accumulated enough learned concepts \cite{bloom2002children}. In particular, it is known that children can form quick and rough hypotheses about the meaning of new words in a sentence based on their knowledge of previous learned words \cite{carey1978acquiring,heibeck1987word}, associate these words to the objects or their properties, and describe novel concepts using sentences with the new words \cite{bloom2002children}.
This phenomenon has been researched for over 30 years by the psychologists and linguists who study the process of word learning \cite{swingley2010fast}.

For the computer vision field, several methods are proposed \cite{fei2006one,salakhutdinov2010one,tommasi2014learning,lazaridou2014wampimuk} to handle the problem of learning new categories of objects from a handful of examples. This task is important in practice because we sometimes do not have enough data for novel concepts and hence need to transfer knowledge from previously learned categories. Moreover, we do not want to retrain the whole model every time we add a few images with novel concepts, especially when the amount of data or model parameters is very big.

\begin{figure}[!tb]
\begin{center}
\includegraphics[width=0.95\linewidth]{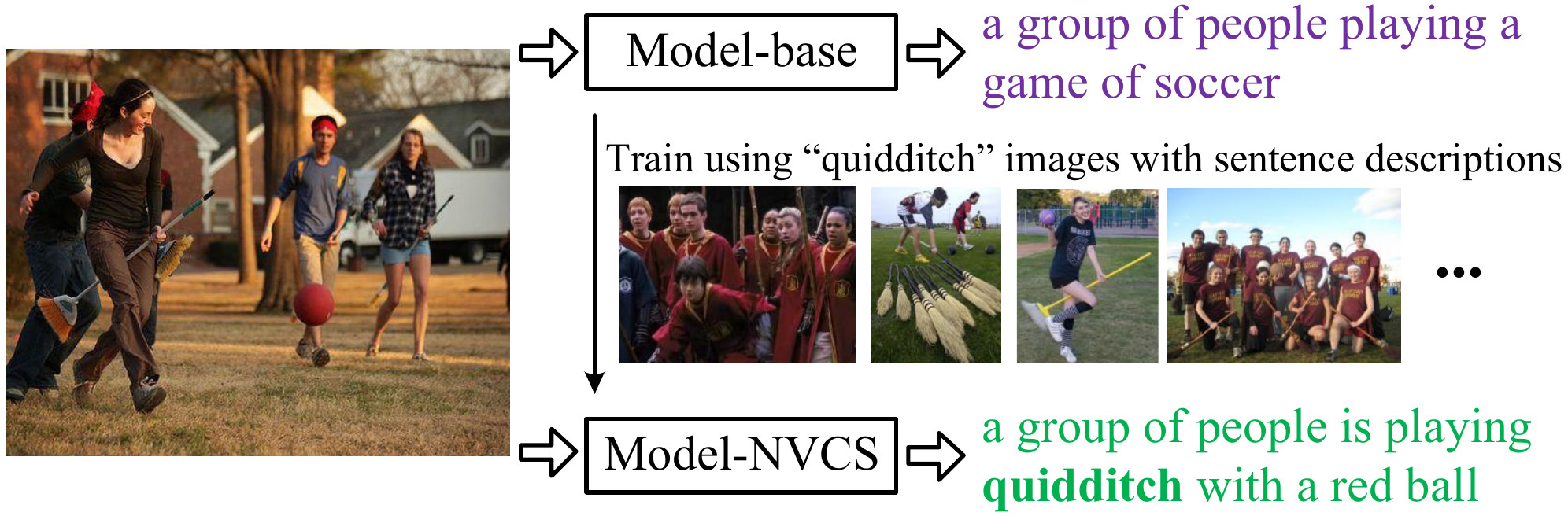}
\end{center}
\vspace{-0.4cm}
   \caption{An illustration of the Novel Visual Concept learning from Sentences (NVCS) task.
   			We start with a model (i.e. model-base) trained with images that do not contain the concept of ``quidditch'' \protect\footnotemark.
   			Using a few ``quidditch'' images with sentence descriptions, our method is able to learn that ``quidditch'' is played by people with a ball.
   		   }
\label{fig:illu_intro}
\vspace{-0.5cm}
\end{figure}

However, these previous methods concentrate on learning classifiers, or mappings, between single words (e.g. a novel object category) and images.  We are unaware of any computer vision studies into the task of learning novel visual concepts from a few sentences and then using these concepts to describe new images -- a task that children seem to do effortlessly. We call this the {\it Novel Visual Concept learning from Sentences (NVCS)} task (see Figure \ref{fig:illu_intro}).

\footnotetext{``quidditch'' is a sport created in ``Harry Potter''. It is played by teams of people holding brooms with a ball (see Figure \ref{fig:illu_intro}).}

In this paper, we present a novel framework to address the NVCS task.
We start with a model that has already been  trained with a large amount of visual concepts.
We propose a method that allows the model to enlarge its word dictionary to describe the novel concepts using a few examples and without extensive retraining.
In particular, we do not need to retrain models from scratch on all of the data (all the previously learned concepts and the novel concepts).
We propose three datasets for the NVCS task to validate our model, which are available on the project page.

Our method requires a {\it base model} for image captioning which will be adapted to perform the NVCS task.
We choose the m-RNN model \cite{mao2014deep}, which performs at the state of the art, as our base model.
Note that we could use most of the current image captioning models as the base model in our method.
But we make several changes to the model structure of m-RNN partly motivated by the desire to avoid overfitting, which is a particular danger for NVCS because we want to learn from a few new images. 
We note that these changes also improve performance on the original image captioning task, although this improvement is not the main focus of this paper. 
In particular, we introduce a transposed weight sharing (TWS) strategy (motivated by auto-encoders \cite{bengio2009learning}) which reduces, by a factor of one half, the number of model  parameters that need to be learned.
This allows us to increase the dimension of the word-embedding and multimodal layers, without overfitting the data, yielding a richer word and multimodal dense representation.
We train this image captioning model on a large image dataset with sentence descriptions. This is the base model which we adapt for the NVCS task. 

Now we address the task of learning the new concepts from a small new set of data that contains these concepts.
There are two main difficulties.
Firstly, the weights for the previously learned concepts may be disturbed by the new concepts. Although this can be solved by fixing these weights. Secondly, learning the new concepts from positive examples can introduce bias.
Intuitively, the model will assign a baseline probability for each word, which is roughly proportional to the frequency of the words in the sentences.
When we train the model on new data, the baseline probabilities of the new words will be unreliably high.
We propose a strategy that addresses this problem by fixing the baseline probability of the new words.

We construct three datasets to validate our method, which involves new concepts of man-made objects, animals, and activities.
The first two datasets are derived from the MS-COCO dataset \cite{lin2014microsoft}.
The third new dataset is constructed by adding three uncommon concepts which do not occur in MS-COCO or other standard datasets. 
These concepts are: quidditch, t-rex and samisen (see section \ref{sec:exp_set})\footnote{The dataset is available at \url{www.stat.ucla.edu/~junhua.mao/projects/child_learning.html}. We are adding more novel concepts in this dataset. The latest version of the dataset contains 8 additional novel concepts: tai-ji, huangmei opera, kiss, rocket gun, tempura, waterfall, wedding dress, and windmill.}.
The experiments show that training our method on only a few examples of the new concepts gives us as good performance as retraining the entire model on all the examples.

\vspace{-0.2cm}
\section{Related Work}
\vspace{-0.2cm}
\begin{figure*}[tbh]
\begin{center}
\includegraphics[width=0.8\linewidth]{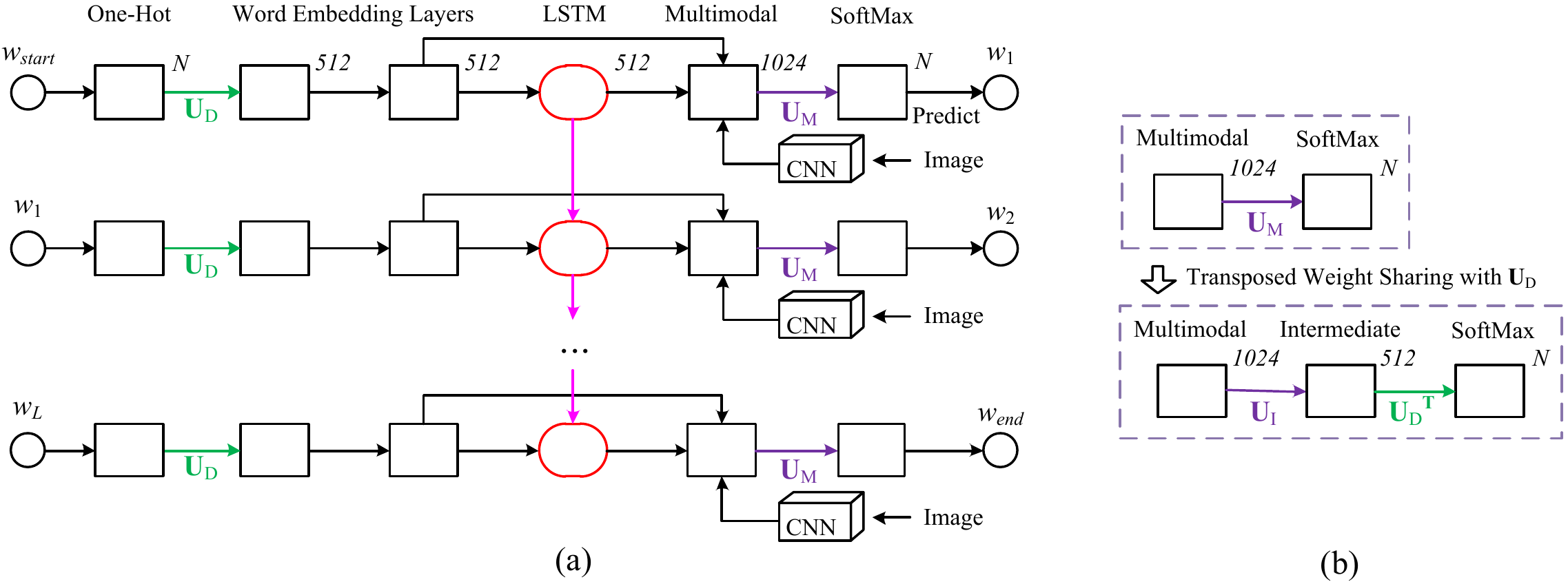}
\end{center}
\vspace{-0.6cm}
   \caption{
   			(a). Our image captioning model.
   			For each word in a sentence, the model takes the current word index and the image as inputs, and outputs the next word index.
   			The weights are shared across the sub-models for the words in a sentence.
   			The number on the top right of each layer denotes its dimension.
   			As in the m-RNN model \cite{mao2014deep}, we add a start sign $w_{start}$ and an end sign $w_{end}$ to each training sentence.
   			(b). The transposed weight sharing of $\mathbf{U}_\mathrm{D}$ and $\mathbf{U}_\mathrm{M}$.
   			(Best viewed in color)
   		   }
\vspace{-0.4cm}
\label{fig:arch_model}
\end{figure*}

\noindent \textbf{Deep neural network}
Recently there have been dramatic progress in deep neural networks for natural language and computer vision.
For natural language, Recurrent Neural Networks (RNNs \cite{elman1990finding,mikolov2010recurrent}) and Long-Short Term Memories (LSTMs \cite{hochreiter1997long}) achieve the state-of-the-art performance for many NLP tasks such as machine translation \cite{kalchbrenner2013recurrent,cho2014learning,sutskever2014sequence} and speech recognition \cite{mikolov2010recurrent}.
For computer vision, deep Convolutional Neural Networks (CNN \cite{lecun2012efficient}) outperform previous methods by a large margin for the tasks of object classification \cite{krizhevsky2012imagenet,simonyan2014very} and detection \cite{girshick2014rcnn,ouyang2014deepid,zhu2014learning}.
The success of these methods for language and vision motivate their use for multimodal learning tasks (e.g. image captioning and sentence-image retrieval).

\noindent \textbf{Multimodal learning of language and vision}
The methods of image-sentence retrieval \cite{frome2013devise,socher2014grounded}, image description generation \cite{kulkarni2011baby,mitchell2012midge,gupta2012image} and visual question-answering \cite{gao2015you,malinowski2014multi,antol2015vqa} have developed very fast in recent years.
Very recent works of image captioning includes  \cite{mao2014explain,kiros2014unifying,karpathy2014deep,vinyals2014show,donahue2014long,fang2014captions,
chen2014learning,lebret2014simple,malinowski2014pooling,klein2014fisher,xu2015show,ma2015multimodal}.
Many of them (e.g. \cite{mao2014deep,vinyals2014show}) adopt an RNN-CNN framework that optimizes the log-likelihood of the caption given the image, and train the networks in an end-to-end way.
An exception is \cite{fang2014captions}, which incorporates visual detectors, language
models, and multimodal similarity models in a high-performing pipeline.
The evaluation metrics of the image captioning task is also discussed \cite{elliott2014comparing,vedantam2014cider}.
All of these image captioning methods use a pre-specified and fixed word dictionary, and train their model on a large dataset.
Our method can be directly applied to any captioning models that adopt an RNN-CNN framework, and our strategy to avoid overfitting is useful for most of the models in the novel visual concept learning task.

\noindent \textbf{Zero-shot and one-shot learning}
For zero-shot learning, the task is to associate dense word vectors or attributes with image features \cite{socher2013zero,frome2013devise,elhoseiny2013write,antol2014zero,lazaridou2014wampimuk}.
The dense word vectors in these papers are pre-trained from a large amount of text corpus and the word semantic representation is captured from co-occurrence with other words \cite{mikolov2013distributed}.
\cite{lazaridou2014wampimuk} developed this idea by only showing the novel words a few times.
In addition, \cite{sharmanska2012augmented} adopted auto-encoders with attribute representations to learn new class labels and \cite{weston2010large} proposed a method that scales to large datasets using label embeddings.

Another related task is one-shot learning task of new categories \cite{fei2006one,lake2011one,tommasi2014learning}.
They learn new objects from only a few examples.
However, these work only consider words or attributes instead of sentences and so their learning target is different from that of the task in this paper.

\vspace{-0.2cm}
\section{The Image Captioning Model}
\vspace{-0.2cm}
We need an image captioning as the base model which will be adapted in the NVCS task.
The base model is based on the m-RNN model \cite{mao2014deep}.
Its architecture is shown in Figure \ref{fig:arch_model}(a).
We make two main modifications of the architecture to make it more suitable for the NVCS task which, as a side effect, also improves performance on the original image captioning task.
Firstly and most importantly, we propose a transposed weight sharing strategy which significantly reduces the number of parameters in the model (see section \ref{sec:model_share}).
Secondly, we replace the recurrent layer in \cite{mao2014deep} by a Long-Short Term Memory (LSTM) layer \cite{hochreiter1997long}.
LSTM is a recurrent neural network which is designed to solve the gradient explosion and vanishing problems.
We briefly introduce the framework of the model in section \ref{sec:model_arch} and describe the details of the transposed weight sharing strategy in section \ref{sec:model_share}.

\vspace{-0.2cm}
\subsection{The Model Architecture}
\label{sec:model_arch}
\vspace{-0.2cm}
As shown in Figure \ref{fig:arch_model}(a), the input of our model for each word in a sentence is the index of the current word in the word dictionary as well as the image.
We represent this index as a one-hot vector (a binary vector with only one non-zero element indicating the index).
The output is the index of the next word.
The model has three components: the language component, the vision component and the multimodal component.
The language component contains two word embedding layers and a LSTM layer.
It maps the index of the word in the dictionary into a semantic dense word embedding space and stores the word context information in the LSTM layer.
The vision component contains a 16-layer deep convolutional neural network (CNN \cite{simonyan2014very}) pre-trained on the ImageNet classification task \cite{ILSVRCarxiv14}.
We remove the final SoftMax layer of the deep CNN and connect the top fully connected layer (a 4096 dimensional layer) to our model.
The activation of this 4096 dimensional layer can be treated as image features that contain rich visual attributes for objects and scenes.
The multimodal component contains a one-layer representation where the information from the language part and the vision part merge together.
We build a SoftMax layer after the multimodal layer to predict the index of the next word.
The weights are shared across the sub-models of the words in a sentence.
As in the m-RNN model \cite{mao2014deep}, we add a start sign $w_{start}$ and an end sign $w_{end}$ to each training sentence.
In the testing stage for image captioning, we input the start sign $w_{start}$ into the model and pick the $K$ best words with maximum probabilities according to the SoftMax layer.
We repeat the process until the model generates the end sign $w_{end}$. 

\vspace{-0.1cm}
\subsection{The Transposed Weight Sharing (TWS)}
\label{sec:model_share}
\vspace{-0.1cm}
For the original m-RNN model \cite{mao2014deep}, most of the weights (i.e. 98.49\%) are contained in the following two weight matrices: $\mathbf{U}_\mathrm{D} \in \mathbbm{R}^{512 \times N}$ and $\mathbf{U}_\mathrm{M} \in \mathbbm{R}^{N \times 1024}$ where $N$ represents the size of the word dictionary.

The weight matrix $\mathbf{U}_\mathrm{D}$ between the one-hot layer and first word embedding layer is used to compute the input of the first word embedding layer $\mathbf{w}(t)$: \vspace{-0.2cm}
\begin{equation}
\mathbf{w}(t)=f(\mathbf{U}_\mathrm{D} \mathbf{h}(t))
\label{equ:Wmap}
\vspace{-0.2cm}
\end{equation}
where $f(.)$ is an element-wise non-linear function, $\mathbf{h}(t) \in \mathbbm{R}^{N\times1}$ is the one-hot vector of the current word.
Note that it is fast to calculate Equation \ref{equ:Wmap} because there is only one non-zero element in $\mathbf{h}(t)$.
In practice, we do not need to calculate the full matrix multiplication operation since only one column of $\mathbf{U}_\mathrm{D}$ is used for each word in the forward and backward propagation.

The weight matrix $\mathbf{U}_\mathrm{M}$ between the multimodal layer and the SoftMax layer is used to compute the activation of the SoftMax layer $\mathbf{y}(t)$:\vspace{-0.2cm}
\begin{equation}
\mathbf{y}(t)=g(\mathbf{U}_\mathrm{M} \mathbf{m}(t) + \mathbf{b})
\label{equ:ori_Mmap}
\vspace{-0.2cm}
\end{equation}
where $\mathbf{m}(t)$ is the activation of the multimodal layer and $g(.)$ is the SoftMax non-linear function.

Intuitively, the role of the weight matrix $\mathbf{U}_\mathrm{D}$ in Equation \ref{equ:Wmap} is to encode the one-hot vector $\mathbf{h}(t)$ into a dense semantic vector $\mathbf{w}(t)$.
The role of the weight matrix $\mathbf{U}_\mathrm{M}$ in Equation \ref{equ:ori_Mmap} is to decode the dense semantic vector $\mathbf{m}(t)$ back to a pseudo one-hot vector $\mathbf{y}(t)$ with the help of the SoftMax function, which is very similar to the inverse operation of Equation \ref{equ:Wmap}.
The difference is that $\mathbf{m}(t)$ is in the dense multimodal semantic space while $\mathbf{w}(t)$ is in the dense word semantic space.

To reduce the number of the parameters, we decompose $\mathbf{U}_\mathrm{M}$ into two parts.
The first part maps the multimodal layer activation vector to an intermediate vector in the word semantic space.
The second part maps the intermediate vector to the pseudo one-hot word vector, which is the inverse operation of Equation \ref{equ:Wmap}.
The sub-matrix of the second part is able to share parameters with $\mathbf{U}_\mathrm{D}$ in a transposed manner, which is motivated by the tied weights strategy in auto-encoders for unsupervised learning tasks \cite{bengio2009learning}.
Here is an example of linear decomposition: $\mathbf{U}_\mathrm{M} = \mathbf{U}_\mathrm{D}^T \mathbf{U}_\mathrm{I}$, where $\mathbf{U}_\mathrm{I} \in \mathbbm{R}^{512 \times 1024}$. 
Equation \ref{equ:ori_Mmap} is accordingly changed to:\vspace{-0.2cm}
\begin{equation}
\vspace{-0.2cm}
\mathbf{y}(t)=g[\mathbf{U}_\mathrm{D}^T \ f(\mathbf{U}_\mathrm{I} \mathbf{m}(t)) + \mathbf{b}]
\label{equ:shared_Mmap}
\end{equation}
where $f(.)$ is a element-wise function.
If $f(.)$ is an identity mapping function, it is equivalent to linearly decomposing $\mathbf{U}_\mathrm{M}$ into $\mathbf{U}_\mathrm{D}^T$ and $\mathbf{U}_\mathrm{I}$.
In our experiments, we find that setting $f(.)$ as the scaled hyperbolic tangent function leads to a slightly better performance than linear decomposition.
This strategy can be viewed as adding an intermediate layer with dimension 512 between the multimodal and SoftMax layers as shown in Figure \ref{fig:arch_model}(b).
The weight matrix between the intermediate and the SoftMax layer is shared with $\mathbf{U}_\mathrm{D}$ in a transposed manner.
This Transposed Weight Sharing (TWS) strategy enables us to use a much larger dimensional word-embedding layer than the m-RNN model \cite{mao2014deep} without increasing the number of parameters.
We also benefit from this strategy when addressing the novel concept learning task.

\begin{figure}[t!]
\begin{center}
\includegraphics[width=0.9\linewidth]{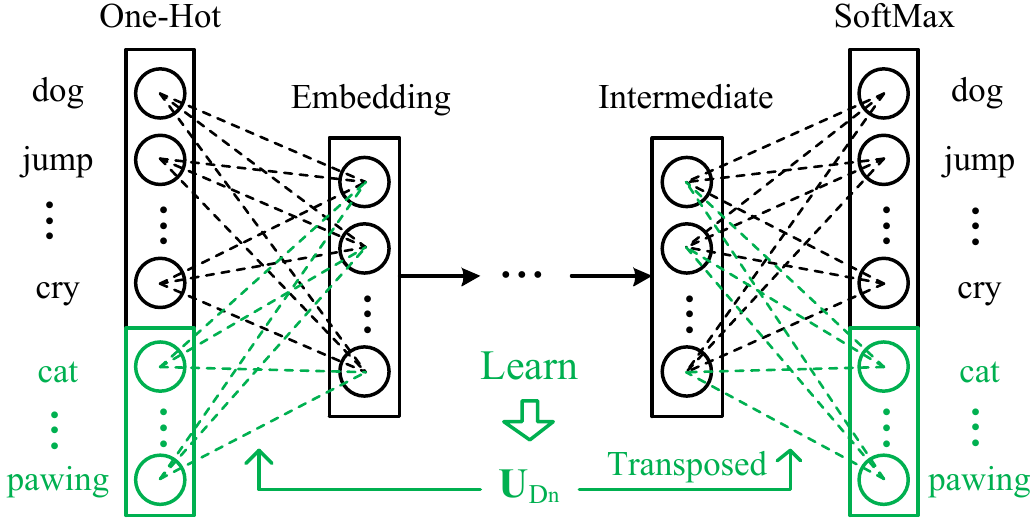}
\end{center}
\vspace{-0.5cm}
   \caption{
   			Illustration of training novel concepts.
   			We only update the sub-matrix $\mathbf{U}_\mathrm{D_n}$ (the green parts in the figure) in $\mathbf{U}_\mathrm{D}$ that is connected to the node of new words in the One-Hot layer and the SoftMax layer during the training for novel concepts.
   			(Best viewed in color)
   		   }
\vspace{-0.5cm}
\label{fig:fixed_weights}
\end{figure}

\vspace{-0.2cm}
\section{The Novel Concept Learning (NVCS) Task}
\vspace{-0.2cm}
Suppose we have trained a model based on a large amount of images and sentences.
Then we meet with images of novel concepts whose sentence annotations contain words not in our dictionary, what should we do?
It is time-consuming and unnecessary to re-train the whole model from scratch using all the data.
In many cases, we cannot even access the original training data of the model.
But fine-tuning the whole model  using only the new data causes severe overfitting on the new concepts and decrease the performance of the model for the originally trained ones.

To solve these problems, we propose the following strategies that learn the new concepts with a few images without losing the accuracy on the original concepts.

\vspace{-0.1cm}
\subsection{Fixing the originally learned weights}
\vspace{-0.1cm}
Under the assumption that we have learned the weights of the original words from a large amount of data and that the amount of the data for new concepts is relatively small, it is straightforward to fix the originally learned weights of the model during the incremental training.
More specifically, the weight matrix $\mathbf{U}_\mathrm{D}$ can be separated into two parts: $\mathbf{U}_\mathrm{D} = [\mathbf{U}_\mathrm{D_o}, \mathbf{U}_\mathrm{D_n}]$, where $\mathbf{U}_\mathrm{D_o}$ and $\mathbf{U}_\mathrm{D_n}$ associate with the original words and the new words respectively.
E.g., as shown in Figure \ref{fig:fixed_weights}, for the novel visual concept ``cat'', $\mathbf{U}_\mathrm{D_n}$ is associated with 29 new words, such as cat, kitten and pawing.
We fix the sub-matrix $\mathbf{U}_\mathrm{D_o}$ and update the sub-matrix $\mathbf{U}_\mathrm{D_n}$ as illustrated in Figure \ref{fig:fixed_weights}.

\vspace{-0.1cm}
\subsection{Fixing the baseline probability}
\vspace{-0.1cm}
In Equation \ref{equ:shared_Mmap}, there is a bias term $\mathbf{b}$.
Intuitively, each element in $\mathbf{b}$ represents the tendency of the model to output the corresponding word.
We can think of this term as the baseline probability of each word.
Similar to $\mathbf{U}_\mathrm{D}$, $\mathbf{b}$ can be separated into two parts: $\mathbf{b} = [\mathbf{b}_o, \mathbf{b}_n]$, where $\mathbf{b}_o$ and $\mathbf{b}_n$ associate with the original words and the new words respectively.
If we only present the new data to the network, the estimation of $\mathbf{b}_n$ is unreliable.
The network will tend to increase the value of $\mathbf{b}_n$ which causes overfitting to the new data.

The easiest way to solve this problem is to fix $\mathbf{b}_n$ during the training for novel concepts.
But this is not enough.
Because the average activation $\mathbf{\bar{x}}$ of the intermediate layer across all the training samples is not $\mathbf{0}$, the weight matrix $\mathbf{U}_\mathrm{D}$ plays a similar role to $\mathbf{b}$ in changing the baseline probability.
To avoid this problem, we centralize the activation of the intermediate layer $\mathbf{x}$ and turn the original bias term $\mathbf{b}$ into $\mathbf{b'}$ as follows:\vspace{-0.2cm}
\begin{equation}
\mathbf{y}(t)=g[\mathbf{U}_\mathrm{D}^T (\mathbf{x}-\mathbf{\bar{x}}) + \mathbf{b'}]; \ \ \mathbf{b'}_o = \mathbf{b}_o + \mathbf{U}_\mathrm{D_o}^T \mathbf{\bar{x}}
\vspace{-0.2cm}
\label{equ:BPF}
\end{equation}
After that, we set every element in $\mathbf{b}_n'$ to be the average value of the elements in $\mathbf{b}_o'$ and fix $\mathbf{b}_n'$ when we train on the new images.
We call this strategy Baseline Probability Fixation (BPF).

In the experiments, we adopt a stochastic gradient descent algorithm with an initial learning rate of 0.01 and use AdaDelta \cite{zeiler2012adadelta} as the adaptive learning rate algorithm for both the base model and the novel concept model.

\vspace{-0.1cm}
\subsection{The Role of Language and Vision}
\vspace{-0.1cm}
\label{sec:rol_nlp&vision}
In the novel concept learning (NVCS) task, the sentences serve as a weak labeling of the image.
The language part of the model (the word embedding layers and the LSTM layer) hypothesizes the basic properties (e.g. the parts of speech) of the new words and whether the new words are closely related to the content of the image.
It also hypothesizes which words in the original dictionary are semantically and syntactically close to the new words.
For example, suppose the model meets a new image with the sentence description ``A woman is playing with a cat''.
Also suppose there are images in the original data containing sentence description such as ``A man is playing with a dog''. 
Then although the model has not seen the word ``cat'' before, it will hypothesize that the word ``cat'' and ``dog'' are close to each other.

The vision part is pre-trained on the ImageNet classification task \cite{ILSVRCarxiv14} with 1.2 million images and 1,000 categories.
It provides rich visual attributes of the objects and scenes that are useful not only for the 1,000 classification task itself, but also for other vision tasks \cite{donahue2013decaf}.

Combining cues from both language and vision, our model can effectively learn the new concepts using only a few examples as demonstrated in the experiments.

\vspace{-0.2cm}
\section{Datasets}
\vspace{-0.2cm}
\label{sec:exp_set}
\begin{figure}[!tb]
\begin{center}
\includegraphics[width=0.8\linewidth]{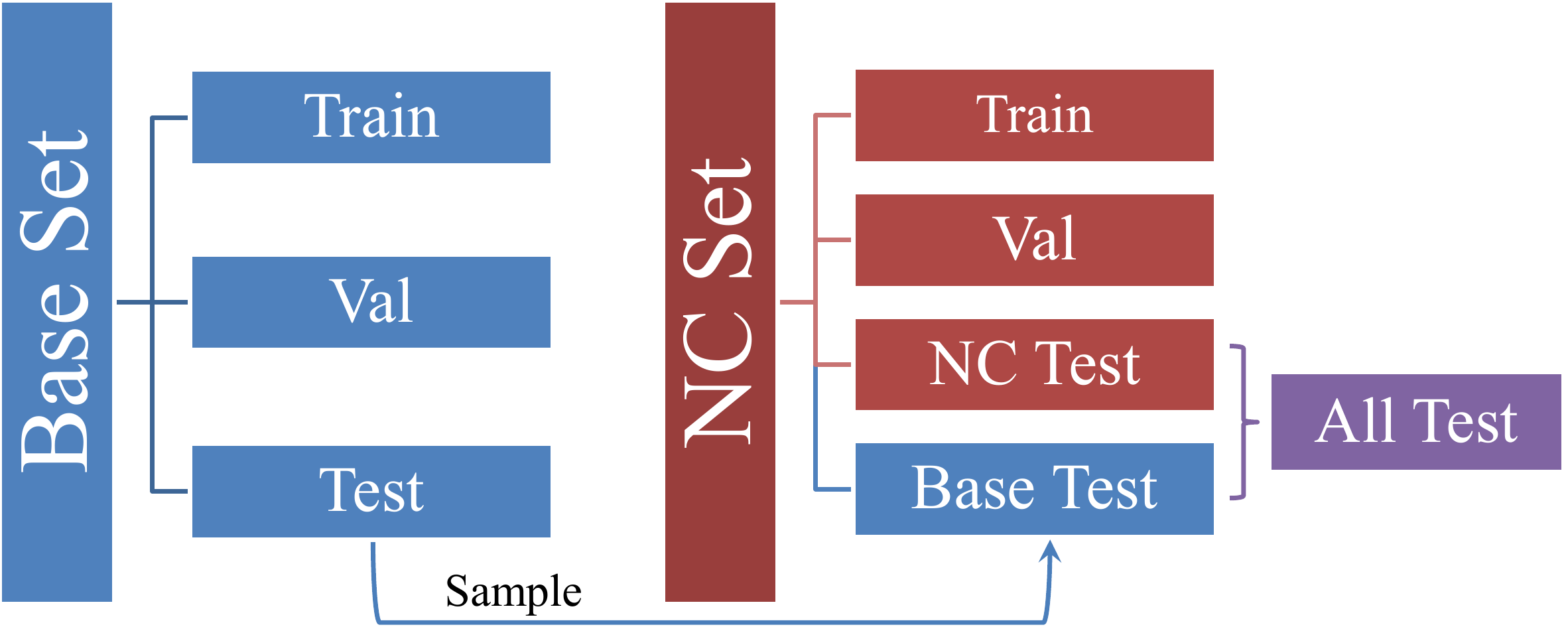}
\end{center}
\vspace{-0.5cm}
   \caption{Organization of the novel concept datasets}
\label{fig:dataset_org}
\vspace{-0.5cm}
\end{figure}

\subsection{Strategies to Construct Datasets}
\vspace{-0.1cm}
We use the annotations and images from the MS COCO \cite{lin2014microsoft} to construct our Novel Concept (NC) learning datasets.
The current release of COCO contains 82,783 training images and 40,504 validation images, with object instance annotations and 5 sentence descriptions for each image.
To construct the NC dataset with a specific new concept (e.g. ``cat''), we remove all images containing the object ``cat'' according to the object annotations.
We also check whether there are some images left with sentences descriptions containing cat related words.
The remaining images are treated as the \emph{Base Set} where we will train, validate and test our base model.
The removed images are used to construct the \emph{Novel Concept set (NC set)}, which is used to train, validate and test our model for the task of novel concept learning.

\vspace{-0.1cm}
\subsection{The Novel Visual Concepts Datasets}
\vspace{-0.1cm}
\label{sec:exp_newcon_set}
\begin{table}[!b]
	\centering
	\tabcolsep=0.1cm
	\renewcommand{\arraystretch}{1.2}
\vspace{-0.4cm}
{\scriptsize
\begin{tabular}{l|ccc}
\noalign{\hrule height 0.8pt}
      & Train & NC Test  & Validation \\
\hline
NewObj-Cat & 2840  & 1000  & 490 \\
NewObj-Motor & 1854  & 600  & 349 \\
NC-3  & 150 ($50 \times 3$) & 120 ($40 \times 3$) & 30 ($10 \times 3$) \\
\noalign{\hrule height 0.8pt}
\end{tabular}%
}
\vspace{-0.2cm}
	\caption{The number of images for the three datasets.}
	\label{tab:dataset_numTr}
\end{table}

We construct three datasets involving five different novel visual concepts:

\textbf{NewObj-Cat} and \textbf{NewObj-Motor}
The corresponding new concepts of these two datasets are \emph{``cat''} and \emph{``motorcycle''} respectively. 
The model need to learn all the related words that describe these concepts and their activities.

\begin{figure}[!tb]
\begin{center}
\includegraphics[width=0.95\linewidth]{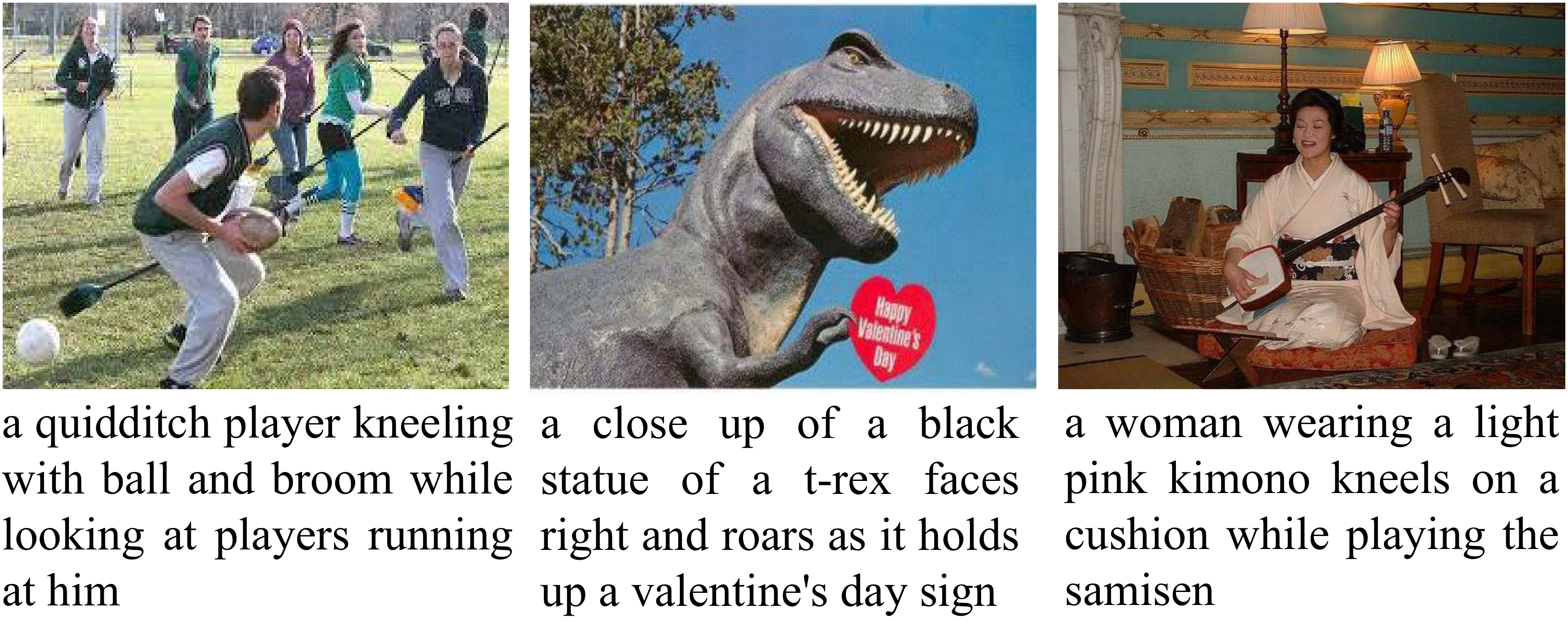}
\end{center}
\vspace{-0.5cm}
   \caption{Sample images and annotations from Novel Concept-3 (NC-3) dataset (see more on the project page).}
\label{fig:NC-3}
\vspace{-0.5cm}
\end{figure}

\textbf{NC-3 dataset}\footnote{The dataset is publicly available at \url{www.stat.ucla.edu/~junhua.mao/projects/child_learning.html}. We are actively expanding the dataset. The latest version contains 11 novel concepts.}
The two datasets mentioned above are all derived from the MS COCO dataset.
To further verify the effectiveness of our method, we construct a new dataset contains three novel concepts: \emph{``quidditch''} (a recently created sport derived from ``Harry Potter''), \emph{``t-rex''} (a dinosaur), and \emph{``samisen''} (an instrument).
It contains not only object concepts (e.g. t-rex and samisen), but also activity concepts (e.g. quidditch).
We labeled 100 images for each concept with 5 sentence annotations for each image.
To diversify the labeled sentences for different images in the same category, the annotators are instructed to label the images with different sentences by describing the details in each image.
It leads to a different style of annotation from that of the MS COCO dataset.
The average length of the sentences is also 26\% longer than that of the MS COCO (13.5 v.s. 10.7).
We construct this dataset for two reasons.
Firstly, the three concepts are not included in the 1,000 categories of the ImageNet Classification task \cite{ILSVRCarxiv14} where we pre-trained the vision component of our model.
Secondly, this dataset has richer and more diversified sentence descriptions compared to NewObj-Cat and NewObj-Motor.
We denote this dataset as Novel Concept-3 dataset (NC-3).
Some samples images and annotations are shown in Figure \ref{fig:NC-3}.

We randomly separate the above three datasets into training, testing and validation sets.
The number of images for the three datasets are shown in Table \ref{tab:dataset_numTr}.
To investigate the possible overfitting issues on these datasets, in the testing stage, we randomly picked images from the testing set of the Base Set and treated them as a separate set of testing images.
The number of added images is equal to the size of the original test set (e.g. 1000 images are picked for NewObj-Cat testing set).
We denote the original new concept testing images as \emph{Novel Concept (NC) test} set and the added base testing images as \emph{Base test} set.
A good novel visual concept learning method should perform better than the base model on NC test set and comparable on Base test set.
The organization of NC datasets is illustrated in Figure \ref{fig:dataset_org}.


\vspace{-0.2cm}
\section{Experiments}
\vspace{-0.2cm}
\subsection{Evaluation Metrics}
\label{sec:sep_eva}
\vspace{-0.1cm}
To evaluate the output sentence descriptions for novel visual concepts, we adopt two evaluation metrics that are widely used in recent image captioning work: BLEU scores \cite{papineni2002bleu} (BLEU score for n-gram is denoted as as B-n in the paper) and METEOR \cite{lavie2007meteor}.


Both BLEU scores and METEOR target on evaluating the overall quality of the generated sentences.
In the NVCS task, however, we focus more on the accuracy for the new words than the previously learned words in the sentences.
Therefore,  to conduct a comprehensive evaluation, we also calculate the $f$ score for the words that describe the new concepts.
E.g. for the cat dataset, there are 29 new words such as cat, cats, kitten, and pawing.
The precision $p$ and recall $r$ for each new word in the dictionary ($w_n^d$) are calculated as follows:

\vspace{-0.4cm}
\begin{scriptsize}
$$
p = \frac{\mathbf{N}(w_n^d \in \mathbb{S}_{gen} \wedge w_n^d \in \mathbb{S}_{ref} )}{\mathbf{N}(w_n^d \in \mathbb{S}_{gen})}; 
r = \frac{\mathbf{N}(w_n^d \in \mathbb{S}_{gen} \wedge w_n^d \in \mathbb{S}_{ref} )}{\mathbf{N}(w_n^d \in \mathbb{S}_{ref})}
$$
\end{scriptsize}
\vspace{-0.4cm}

\noindent where $\mathbb{S}_{gen}$ denotes generated sentence, $\mathbb{S}_{ref}$ denotes reference sentences, $\mathbf{N}(condition)$ represents number of testing images that conform to the condition.
Note that $p$ and $r$ are calculated on the combined testing set of the NC test set and the base test set (i.e. All test).

A high $r$ with a low $p$ indicates that the model overfits the new data (We can always get $r=1$ if we output the new word every time) while a high $p$ with a low $r$ indicates underfiting.
We use the $f = \frac{2}{p^{-1} + r^{-1}}$ as a balanced measurement between $p$ and $r$.
Best $f$ score is 1.
Note that $f=0$ if either $p=0$ or $r=0$.
Compared to METEOR and BLEU, the $f$ score show the effectiveness of the model to learn new concepts more explicitly.

\vspace{-0.1cm}
\subsection{Effectiveness of TWS and BPF}
\label{sec:sep_eff}
\vspace{-0.1cm}

\begin{table}[!tb]
	\centering
	\tabcolsep=0.1cm
	\renewcommand{\arraystretch}{1.2}
{\scriptsize
\begin{tabular}{l|ccccccc}
\noalign{\hrule height 0.8pt}
      & B-1   & B-2   & B-3   & B-4   & METEOR & CIDEr & ROUGE\_L \\
\hline
m-RNN \cite{mao2014deep} & 0.680 & 0.506 & 0.369 & 0.272 & 0.225 & 0.791 & 0.499 \\
ours-TWS & \textbf{0.685} & \textbf{0.512} & \textbf{0.376} & \textbf{0.279} & \textbf{0.229} & \textbf{0.819} & \textbf{0.504} \\
\noalign{\hrule height 0.8pt}
\end{tabular}}
\vspace{-0.2cm}
	\caption{The performance comparisons of our model and m-RNN \cite{mao2014deep} for the standard image captioning task.}
	\label{tab:res_imagecaption}
\vspace{-0.5cm}
\end{table}

We test our base model with the Transposed Weight Sharing (TWS) strategy in the original image captaining task on the MS COCO \cite{capeval2015} and compare to m-RNN \cite{mao2014deep}, which does not use TWS. 
Our model performs better than m-RNN in this task as shown in Table \ref{tab:res_imagecaption}.
We choose the layer dimensions of our model so that the number of parameters matches that of \cite{mao2014deep}.
Models with different hyper-parameters, features or pipelines might lead to better performance, which is beyond the scope of this paper. E.g. \cite{mao2014deep,vinyals2014show,fang2014captions} further improve their results after the submission of this draft and achieve a B-4 score of 0.302, 0.309 and 0.308 respectively using, e.g., fine-tuned image features on COCO or consensus reranking \cite{devlin2015exploring,mao2014deep}, which are complementary with TWS.

We also validate the effectiveness of our Transposed Weight Sharing (TWS) and Baseline Probability Fixation (BPF) strategies for the novel concept learning task on the NewObj-Cat dataset.
We compare the performance of five Deep-NVCS models.
Their properties and performance in terms of $f$ score for the word ``cat'' are summarized in Table \ref{tab:res_BPF_TWS}.
``BiasFix'' means that we fix the bias term $\mathbf{b}_n$ in Equation \ref{equ:shared_Mmap}.
``Centralize'' means that we centralize the intermediate layer activation $\mathbf{x}$ (see Equation \ref{equ:BPF}) so that $\mathbf{U}_\mathrm{D}$ will not affect the baseline probability.

We achieve 2.5\% increase of performance in terms of $f$ using TWS (Deep-NVCS-BPF-TWS v.s. Deep-NVCS-BPF-noTWS\footnote{We tried two versions of the model without TWS: (\rom{1}). the model with multimodal layer directly connected to softmax layer like \cite{mao2014deep}, (\rom{2}). the model with an additional intermediate layer like TWS but does not share the weights. In our experiments, (\rom{1}) performs slightly better than (\rom{2}) so we report the performance of (\rom{1}) here.}), and achieves 2.4\% increase using BPF (Deep-NVCS-BPF-TWS v.s. Deep-NVCS-UnfixedBias). 
We use Deep-NVCS to represent Deep-NVCS-BPF-TWS in short for the rest of the paper.

\begin{table}[!tb]
	\centering
	\renewcommand{\arraystretch}{1.2}
{\scriptsize
\begin{tabular}{l|ccc|c}
\noalign{\hrule height 0.8pt}
      & BiasFix & Centralize   & TWS   & $f$ \\
\hline
Deep-NVCS-UnfixedBias & $\times$     & $\times$     & $\surd$     & 0.851 \\
Deep-NVCS-FixedBias & $\surd$     & $\times$     & $\surd$     & 0.860 \\
Deep-NVCS-NoBPF-NoTWS & $\times$     & $\times$     & $\times$     & 0.839 \\
Deep-NVCS-BPF-NoTWS & $\surd$     & $\surd$     & $\times$     & 0.850 \\
Deep-NVCS-BPF-TWS & $\surd$     & $\surd$     & $\surd$     & \textbf{0.875} \\
\noalign{\hrule height 0.8pt}
\end{tabular}}
\vspace{-0.2cm}
	\caption{Performance of Deep-NVCS models with different novel concept learning strategies on NewObj-Cat.
			 TWS and BPF improve the performance.
			}
	\label{tab:res_BPF_TWS}
\vspace{-0.5cm}
\end{table}

\vspace{-0.1cm}
\subsection{Results on NewObj-Motor and NewObj-Cat}
\label{sec:exp_newobj_cat}
\begin{table*}[!tb]
	\centering
	\tabcolsep=0.15cm
	\renewcommand{\arraystretch}{1.2}
{\scriptsize
\begin{tabular}{l|c|ccccc|ccccc|ccccc}
\noalign{\hrule height 0.8pt}
      &  & \multicolumn{5}{c|}{All test}       & \multicolumn{5}{c|}{NC test} & \multicolumn{5}{c}{Base test} \\
Evaluation Metrics  & \textit{f} & B-1   & B-2   & B-3   & B-4   & {\tiny METEOR} & B-1   & B-2   & B-3   & B-4   & {\tiny METEOR} & B-1   & B-2 & B-3 & B-4 & {\tiny METEOR} \\
\hline
\multicolumn{17}{c}{NewObj-Cat}\\
\hline
Model-retrain & 0.866 & 0.689 & 0.531 & 0.391 & 0.291 & 0.228 & 0.696 & 0.549 & 0.403 & 0.305 & 0.227 & 0.683 & 0.513 & 0.379 & 0.277 & 0.229 \\
Model-base & 0.000     & 0.645 & 0.474 & 0.339 & 0.247 & 0.201 & 0.607 & 0.436 & 0.303 & 0.217 & 0.175 & 0.683 & 0.511 & 0.375 & 0.277 & 0.227 \\
Model-word2vec & 0.183 & 0.642 & 0.471 & 0.341 & 0.245 & 0.200 & 0.610 & 0.432 & 0.307 & 0.217 & 0.176 & 0.674 & 0.510 & 0.375 & 0.273 & 0.224 \\
Deep-NVCS & 0.875 & 0.682 & 0.521 & 0.382 & 0.286 & 0.224 & 0.684 & 0.534 & 0.392 & 0.299 & 0.224 & 0.680 & 0.508 & 0.372 & 0.274 & 0.225 \\
Deep-NVCS-1:1Inc & 0.881 & 0.683 & 0.523 & 0.385 & 0.288 & 0.226 & 0.686 & 0.538 & 0.398 & 0.303 & 0.226 & 0.679 & 0.507 & 0.371 & 0.273 & 0.225 \\
\hline
\multicolumn{17}{c}{NewObj-Motor}\\
\hline
Model-retrain & 0.797 & 0.697 & 0.526 & 0.386 & 0.284 & 0.240 & 0.687 & 0.512 & 0.368 & 0.263 & 0.244 & 0.707 & 0.539 & 0.404 & 0.305 & 0.236 \\
Model-base & 0.000    & 0.646 & 0.460 & 0.327 & 0.235 & 0.218 & 0.586 & 0.380 & 0.245 & 0.160 & 0.203 & 0.705 & 0.536 & 0.401 & 0.301 & 0.235 \\
Model-word2vec & 0.279 & 0.663 & 0.476 & 0.338 & 0.243 & 0.226 & 0.624 & 0.423 & 0.279 & 0.183 & 0.223 & 0.701 & 0.530 & 0.397 & 0.303 & 0.229 \\
Deep-NVCS & 0.784 & 0.688 & 0.512 & 0.373 & 0.276 & 0.236 & 0.677 & 0.494 & 0.349 & 0.252 & 0.241 & 0.698 & 0.530 & 0.398 & 0.299 & 0.231 \\
Deep-NVCS-1:1Inc & 0.790 & 0.687 & 0.512 & 0.374 & 0.280 & 0.235 & 0.672 & 0.492 & 0.347 & 0.256 & 0.237 & 0.702 & 0.532 & 0.401 & 0.303 & 0.234 \\
\noalign{\hrule height 0.8pt}
\end{tabular}%
}
\vspace{-0.2cm}
	\caption{
			 Results on the NewObj-Cat and NewObj-Motor dataset using all the training samples.
			 The Deep NVCS models outperform the simple baselines. They achieve comparable performance with the strong baseline (i.e. Model-retrain) but only need $\leq$2\% of the time.
			 Model-base and Model-retrain stand for the model trained on base set (no novel concepts) and the model retrained on the combined data (all the images of base set and novel concept set) respectively.
			 Model-word2vec is a baseline model based on word2vec \cite{mikolov2013distributed}.
			 Deep-NVCS stands for the model trained only with the new concept data.
			 Deep-NVCS-1:1Inc stands for the Deep-NVCS model trained by adding equal number of training images from the base set.
			}
	\label{tab:res_all_motor_cat_B}
\vspace{-0.2cm}
\end{table*}

\begin{figure*}[!tb]
\centering
	\begin{minipage}[b]{0.95\textwidth}
	\includegraphics[width=1\textwidth]{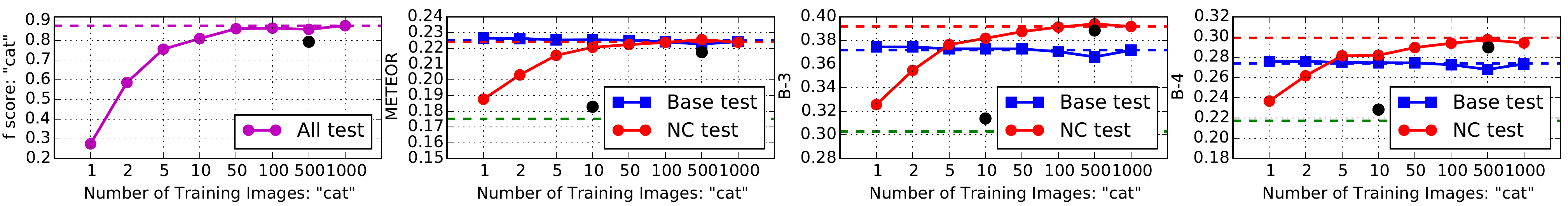} \\
	\includegraphics[width=1\textwidth]{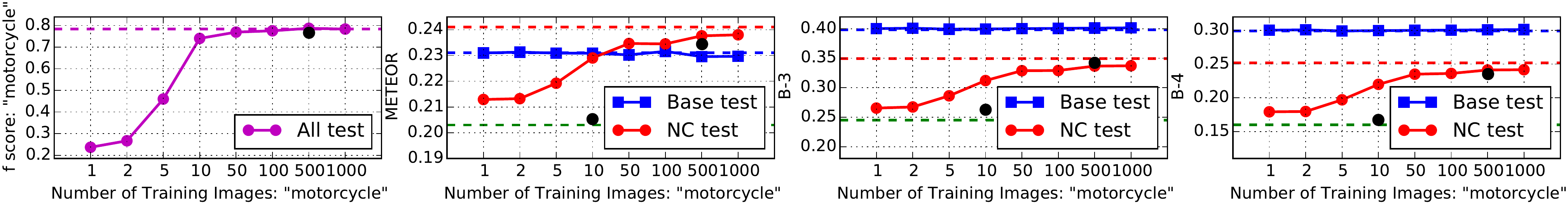}
	\end{minipage}
	\vspace{-0.2cm}
	\caption{
			 Performance comparison of our model with different number of training images on NewObj-cat and NewObj-Motor datasets.
   			 The red, blue, and magenta dashed line indicates the performance of Deep-NVCS using all the training images on the base test set, the NC test set and the all test set respectively.
    		 The green dashed line indicates the performance of Model-base.
    		 The black dots stand for the performance of Model-retrain for NC test.
			 We show that our model trained with 10 to 50 images achieves comparable performance with the model trained on the full training set.
			 (Best viewed in color)
			}
	\vspace{-0.5cm}
	\label{tab:res_os_motor_cat_B}
\end{figure*}

\vspace{-0.1cm}
\subsubsection{Using all training samples}
\vspace{-0.2cm}
We show the performance of our Deep-NVCS models compared to strong baselines on the NewObj-Cat and NewObj-Motor datasets in Table \ref{tab:res_all_motor_cat_B}.
For Deep-NVCS, we only use the training data from the novel concept set.
For Deep-NVCS-Inc1:1, we add training data randomly sampled from the training set of the base set.
The number of added training images is the same as that of the training images for novel concepts.
\textit{Model-base} stands for the model trained only on the base set (no novel concept images).
We implement a baseline model, \textit{Model-word2vec}, where the weights of new words ($\mathbf{U}_\mathrm{D_n}$) are calculated using a weighted sum of the weights of 10 similar concepts measured by the unsupervised learned word-embeddings from word2vec \cite{mikolov2013distributed}.
We also implement a strong baseline, \textit{Model-retrain}, by retraining the whole model from scratch on the combined training set (training images from both the base set and the NC set).

The results show that compared to the Model-base which is only trained on base set, the Deep-NVCS models perform much better on the novel concept test set while reaching comparable performance on the base test set.
Deep-NVCS also performs better than the Model-word2vec model.
The performance of our Deep-NVCS models is very close to that of the strong baseline Model-retrain but needs \textit{only less than 2\% of the time}.
This demonstrates the effectiveness of our novel concept learning strategies.
The model learns the new words for the novel concepts without disturbing the previous learned words.

The performance of Deep-NVCS is also comparable with, though slightly lower than Deep-NVCS-1:1Inc.
Intuitively, if the image features can successfully capture the difference between the new concepts and the existing ones, it is sufficient to learn the new concept only from the new data.
However, if the new concepts are very similar to some previously learned concepts, such as cat and dog, it is helpful to present the data of both novel and existing concepts to make it easier for the model to find the difference.

\begin{figure*}[!tb]
\begin{center}
\includegraphics[width=0.9\linewidth]{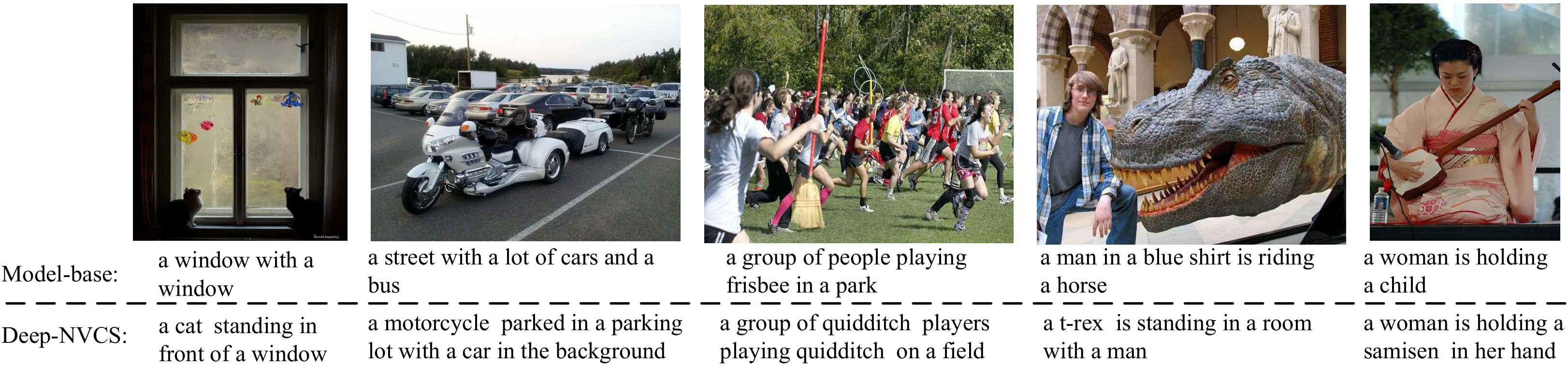}
\end{center}
\vspace{-0.6cm}
   \caption{
   			The generated sentences for the test images from novel concept datasets.
   			In these examples, cat, motorcycle, quidditch, t-rex and samisen are the novel concepts respectively.
   		   }
\vspace{-0.5cm}
\label{fig:sample_res}
\end{figure*}

\vspace{-0.3cm}
\subsubsection{Using a few training samples}
\vspace{-0.1cm}
We also test our model under the one or few-shot scenarios. 
Specifically, we randomly sampled $k$ images from the training set of NewObj-Cat and NewObj-Motor, and trained our Deep-NVCS model only on these images ($k$ ranges from 1 to 1000).
We conduct the experiments 10 times and average the results to avoid the randomness of the sampling.

We show the performance of our model with different number of training images in Figure \ref{tab:res_os_motor_cat_B}.
We only show the results in terms of $f$ score, METEOR, B-3 and B-4 because of space limitation.
The results of B-1 and B-2 and consistent with the shown metrics.
The performance of the model trained with the full NC training set in the last section is indicated by the blue (Base test), red (NC test) or magenta (All test) dashed lines in Figure \ref{tab:res_os_motor_cat_B}.
These lines represent the experimental upper bounds of our model under the one or few-shot scenario.
The performance of the Model-base is shown by a green dashed line.
It serves as an experimental lower bound.
We also show the results of Model-retrain for NC test with black dots in Figure \ref{tab:res_os_motor_cat_B} trained with 10 and 500 novel concepts images.

The results show that using about 10 to 50 training images, the model achieves comparable performance with the Deep-NVCS model trained on the full novel concept training set.
In addition, using about 5 training images, we observe a nontrivial increase of performance compared to the base model.
Our deep-NVCS also better handles the case for a few images and runs much faster than Model-retrain.

\vspace{-0.1cm}
\subsection{Results on NC-3}
\label{sec:exp_nc-3}
\vspace{-0.1cm}

\begin{table}[b!]
	\centering
	\tabcolsep=0.1cm
	\renewcommand{\arraystretch}{1.2}
\vspace{-0.4cm}
{\scriptsize
\begin{tabular}{l|cccc|cccc}
\noalign{\hrule height 0.8pt}
\multicolumn{1}{l}{Evaluation Metrics} & \textit{f}     & B-3   & B-4   & MET. & \textit{f}     & B-3   & B-4   & MET. \\
\noalign{\hrule height 0.8pt}
& \multicolumn{4}{c|}{quidditch}         & \multicolumn{4}{c}{t-rex} \\
\hline
\multicolumn{1}{l|}{Model-retrain} & 0.000 & 0.196 & 0.138 & 0.120 & 0.213 & 0.224 & 0.141 & 0.105 \\
\multicolumn{1}{l|}{Model-base} & 0.000     & 0.193 & 0.139 & 0.122 & 0.000     & 0.166 & 0.102 & 0.088 \\
\multicolumn{1}{l|}{Deep-NVCC} & 0.854 & 0.237 & 0.167 & 0.168 & 0.861 & 0.247 & 0.144 & 0.187 \\
\multicolumn{1}{l|}{Deep-NVCC-1:1Inc} & 0.863 & 0.244 & 0.170 & 0.170 & 0.856 & 0.242 & 0.132 & 0.186 \\
\hline
& \multicolumn{4}{c|}{samisen}           & \multicolumn{4}{c}{Base Test} \\
\hline
\multicolumn{1}{l|}{Model-retrain} & 0.000 & 0.209 & 0.133 & 0.122 & -     & 0.412 & 0.328 & 0.234 \\
\multicolumn{1}{l|}{Model-base} & 0.000    & 0.177 & 0.105 & 0.122 & -     & 0.414 & 0.325 & 0.240 \\
\multicolumn{1}{l|}{Deep-NVCC} & 0.630 & 0.229 & 0.140 & 0.161 & -     & 0.414 & 0.326 & 0.239 \\
\multicolumn{1}{l|}{Deep-NVCC-1:1Inc} & 0.642 & 0.233 & 0.144 & 0.164 & -     & 0.414 & 0.327 & 0.239 \\
\noalign{\hrule height 0.8pt}
\end{tabular}}
\vspace{-0.2cm}
	\caption{
			  Results of our model on the NC-3 Datasets.
			}
	\label{tab:res_NC-3_B}
\end{table}

The NC-3 dataset has three main difficulties.
Firstly, the concepts have very similar counterparts in the original image set, such as Samisen v.s. Guitar, Quidditch v.s. football.
Secondly, the three concepts rarely appear in daily life.
They are not included in the ImageNet 1,000 categories where we pre-trained our vision deep CNN.
Thirdly, the way we describe the three novel concepts is somewhat different from that of the common objects included in the base set.
The requirement to diversify the annotated sentences makes the difference of the style for the annotated sentences between NC-3 and MS COCO even larger.
The effect of the difference in sentence style leads to decreased performance of the base model compared to that on the NewObj-Cat and NewObj-Motor dataset (see Model-base in Table \ref{tab:res_NC-3_B} compared to that in Table \ref{tab:res_all_motor_cat_B} on NC test).
Furthermore, it makes it harder for the model to hypothesize the meanings of new words from a few sentences.

Faced with these difficulties, our model still learns the semantic meaning of the new concepts quite well.
The $f$ scores of the model shown in Table \ref{tab:res_NC-3_B} indicate that the model successfully learns the new concepts with a high accuracy from only 50 examples.

It is interesting that Model-retrain performs very badly on this dataset.
It does not output the word ``quidditch'' and ``samisen'' in the generated sentences.
The BLEU scores and METEOR are also very low.
This is not surprising since there are only a few training examples (i.e. 50) for these three novel concepts and so it is easy to be overwhelmed by other concepts from the original MS COCO dataset.

%

\vspace{-0.1cm}
\subsection{Qualitative Results}
\vspace{-0.1cm}

\begin{table}[!tb]
	\centering
    \tabcolsep=0.3cm
	\renewcommand{\arraystretch}{1.2}
{\scriptsize
\begin{tabular}{l|l}
\noalign{\hrule height 0.8pt}
New Word & Five nearest neighbours \\
\hline
cat   & kitten; tabby; puppy; calico; doll; \\
motorcycle & motorbike; moped; vehicle; motor; motorbikes; \\
quidditch & soccer; football; softball; basketball; frisbees; \\
t-rex & giraffe's; bull; pony; goat; burger; \\
samisen & guitar; wii; toothbrushes; purse; contents; \\
\noalign{\hrule height 0.8pt}
\end{tabular}%
}
	\caption{The five nearest neighbors of the new words as measured by the activation of the word-embedding layer.}
	\label{tab:wv_near}
	\vspace{-0.5cm}
\end{table}


In Table \ref{tab:wv_near}, we show the five nearest neighbors of the new concepts using the activation of the word-embedding layer learned by our Deep-NVCS model.
It shows that the learned novel word embedding vectors captures the semantic information from both language and vision.
We also show some sample generated sentence descriptions of the base model and our Deep-NVCS model in Figure \ref{fig:sample_res}.

\vspace{-0.2cm}
\section{Conclusion}
\vspace{-0.2cm}

In this paper, we propose the Novel Visual Concept learning from Sentences (NVCS) task.
In this task, methods need to learn novel concepts from sentence descriptions of a few images.
We describe a method that allows us to train our model on a small number of images containing novel concepts.
This performs comparably with the model retrained from scratch on all of the data if the number of novel concept images is large, and performs better when there are only a few training images of novel concepts available.
We construct three novel concept datasets where we validate the effectiveness of our method.
These datasets have been released to encourage future research in this area.

\vspace{-0.1cm}
\section*{\normalsize{Acknowledgement}}
\vspace{-0.2cm}
\small{
We thank the comments and suggestions of the anonymous reviewers, and help from Xiaochen Lian in the dataset collection process.
We acknowledge support from NSF STC award CCF-1231216 and ARO 62250-CS.
}

{\small
\bibliographystyle{ieee}
\bibliography{egbib}

\begin{thebibliography}{10}\itemsep=-1pt

\bibitem{antol2015vqa}
S.~Antol, A.~Agrawal, J.~Lu, M.~Mitchell, D.~Batra, C.~L. Zitnick, and
  D.~Parikh.
\newblock Vqa: Visual question answering.
\newblock {\em arXiv}, 2015.

\bibitem{antol2014zero}
S.~Antol, C.~L. Zitnick, and D.~Parikh.
\newblock Zero-shot learning via visual abstraction.
\newblock In {\em ECCV}, pages 401--416. 2014.

\bibitem{bengio2009learning}
Y.~Bengio.
\newblock Learning deep architectures for ai.
\newblock {\em Foundations and trends{\textregistered} in Machine Learning},
  2(1):1--127, 2009.

\bibitem{bloom2002children}
P.~Bloom.
\newblock {\em How children learn the meanings of words}.
\newblock MIT press, 2002.

\bibitem{carey1978acquiring}
S.~Carey and E.~Bartlett.
\newblock Acquiring a single new word.
\newblock 1978.

\bibitem{capeval2015}
X.~Chen, H.~Fang, T.~Lin, R.~Vedantam, S.~Gupta, P.~Dollár, and C.~L. Zitnick.
\newblock Microsoft coco captions: Data collection and evaluation server.
\newblock {\em arXiv preprint arXiv:1504.00325}, 2015.

\bibitem{chen2014learning}
X.~Chen and C.~L. Zitnick.
\newblock Learning a recurrent visual representation for image caption
  generation.
\newblock {\em arXiv preprint arXiv:1411.5654}, 2014.

\bibitem{cho2014learning}
K.~Cho, B.~van Merrienboer, C.~Gulcehre, F.~Bougares, H.~Schwenk, and
  Y.~Bengio.
\newblock Learning phrase representations using rnn encoder-decoder for
  statistical machine translation.
\newblock {\em arXiv preprint arXiv:1406.1078}, 2014.

\bibitem{devlin2015exploring}
J.~Devlin, S.~Gupta, R.~Girshick, M.~Mitchell, and C.~L. Zitnick.
\newblock Exploring nearest neighbor approaches for image captioning.
\newblock {\em arXiv preprint arXiv:1505.04467}, 2015.

\bibitem{donahue2014long}
J.~Donahue, L.~A. Hendricks, S.~Guadarrama, M.~Rohrbach, S.~Venugopalan,
  K.~Saenko, and T.~Darrell.
\newblock Long-term recurrent convolutional networks for visual recognition and
  description.
\newblock {\em arXiv preprint arXiv:1411.4389}, 2014.

\bibitem{donahue2013decaf}
J.~Donahue, Y.~Jia, O.~Vinyals, J.~Hoffman, N.~Zhang, E.~Tzeng, and T.~Darrell.
\newblock Decaf: A deep convolutional activation feature for generic visual
  recognition.
\newblock {\em arXiv preprint arXiv:1310.1531}, 2013.

\bibitem{elhoseiny2013write}
M.~Elhoseiny, B.~Saleh, and A.~Elgammal.
\newblock Write a classifier: Zero-shot learning using purely textual
  descriptions.
\newblock In {\em ICCV}, pages 2584--2591, 2013.

\bibitem{elliott2014comparing}
D.~Elliott and F.~Keller.
\newblock Comparing automatic evaluation measures for image description.
\newblock In {\em ACL}, volume~2, pages 452--457, 2014.

\bibitem{elman1990finding}
J.~L. Elman.
\newblock Finding structure in time.
\newblock {\em Cognitive science}, 14(2):179--211, 1990.

\bibitem{fang2014captions}
H.~Fang, S.~Gupta, F.~Iandola, R.~Srivastava, L.~Deng, P.~Doll{\'a}r, J.~Gao,
  X.~He, M.~Mitchell, J.~Platt, et~al.
\newblock From captions to visual concepts and back.
\newblock {\em arXiv preprint arXiv:1411.4952}, 2014.

\bibitem{fei2006one}
L.~Fei-Fei, R.~Fergus, and P.~Perona.
\newblock One-shot learning of object categories.
\newblock {\em TPAMI}, 28(4):594--611, 2006.

\bibitem{frome2013devise}
A.~Frome, G.~S. Corrado, J.~Shlens, S.~Bengio, J.~Dean, T.~Mikolov, et~al.
\newblock Devise: A deep visual-semantic embedding model.
\newblock In {\em NIPS}, pages 2121--2129, 2013.

\bibitem{gao2015you}
H.~Gao, J.~Mao, J.~Zhou, Z.~Huang, L.~Wang, and W.~Xu.
\newblock Are you talking to a machine? dataset and methods for multilingual
  image question answering.
\newblock In {\em NIPS}, 2015.

\bibitem{girshick2014rcnn}
R.~Girshick, J.~Donahue, T.~Darrell, and J.~Malik.
\newblock Rich feature hierarchies for accurate object detection and semantic
  segmentation.
\newblock In {\em CVPR}, 2014.

\bibitem{gupta2012image}
A.~Gupta and P.~Mannem.
\newblock From image annotation to image description.
\newblock In {\em ICONIP}, 2012.

\bibitem{heibeck1987word}
T.~H. Heibeck and E.~M. Markman.
\newblock Word learning in children: An examination of fast mapping.
\newblock {\em Child development}, pages 1021--1034, 1987.

\bibitem{hochreiter1997long}
S.~Hochreiter and J.~Schmidhuber.
\newblock Long short-term memory.
\newblock {\em Neural computation}, 9(8):1735--1780, 1997.

\bibitem{kalchbrenner2013recurrent}
N.~Kalchbrenner and P.~Blunsom.
\newblock Recurrent continuous translation models.
\newblock In {\em EMNLP}, pages 1700--1709, 2013.

\bibitem{karpathy2014deep}
A.~Karpathy and L.~Fei-Fei.
\newblock Deep visual-semantic alignments for generating image descriptions.
\newblock {\em arXiv preprint arXiv:1412.2306}, 2014.

\bibitem{kiros2014unifying}
R.~Kiros, R.~Salakhutdinov, and R.~S. Zemel.
\newblock Unifying visual-semantic embeddings with multimodal neural language
  models.
\newblock {\em arXiv preprint arXiv:1411.2539}, 2014.

\bibitem{klein2014fisher}
B.~Klein, G.~Lev, G.~Sadeh, and L.~Wolf.
\newblock Fisher vectors derived from hybrid gaussian-laplacian mixture models
  for image annotation.
\newblock {\em arXiv preprint arXiv:1411.7399}, 2014.

\bibitem{krizhevsky2012imagenet}
A.~Krizhevsky, I.~Sutskever, and G.~E. Hinton.
\newblock Imagenet classification with deep convolutional neural networks.
\newblock In {\em NIPS}, pages 1097--1105, 2012.

\bibitem{kulkarni2011baby}
G.~Kulkarni, V.~Premraj, S.~Dhar, S.~Li, Y.~Choi, A.~C. Berg, and T.~L. Berg.
\newblock Baby talk: Understanding and generating image descriptions.
\newblock In {\em CVPR}, 2011.

\bibitem{lake2011one}
B.~M. Lake, R.~Salakhutdinov, J.~Gross, and J.~B. Tenenbaum.
\newblock One shot learning of simple visual concepts.
\newblock In {\em CogSci}, volume 172, 2011.

\bibitem{lavie2007meteor}
A.~Lavie and A.~Agarwal.
\newblock Meteor: An automatic metric for mt evaluation with high levels of
  correlation with human judgements.
\newblock In {\em Workshop on Statistical Machine Translation}, pages 228--231,
  2007.

\bibitem{lazaridou2014wampimuk}
A.~Lazaridou, E.~Bruni, and M.~Baroni.
\newblock Is this a wampimuk? cross-modal mapping between distributional
  semantics and the visual world.
\newblock In {\em ACL}, pages 1403--1414, 2014.

\bibitem{lebret2014simple}
R.~Lebret, P.~O. Pinheiro, and R.~Collobert.
\newblock Simple image description generator via a linear phrase-based
  approach.
\newblock {\em arXiv preprint arXiv:1412.8419}, 2014.

\bibitem{lecun2012efficient}
Y.~A. LeCun, L.~Bottou, G.~B. Orr, and K.-R. M{\"u}ller.
\newblock Efficient backprop.
\newblock In {\em Neural networks: Tricks of the trade}, pages 9--48. Springer,
  2012.

\bibitem{lin2014microsoft}
T.-Y. Lin, M.~Maire, S.~Belongie, J.~Hays, P.~Perona, D.~Ramanan,
  P.~Doll{\'a}r, and C.~L. Zitnick.
\newblock Microsoft coco: Common objects in context.
\newblock {\em arXiv preprint arXiv:1405.0312}, 2014.

\bibitem{ma2015multimodal}
L.~Ma, Z.~Lu, L.~Shang, and H.~Li.
\newblock Multimodal convolutional neural networks for matching image and
  sentence.
\newblock {\em arXiv preprint arXiv:1504.06063}, 2015.

\bibitem{malinowski2014multi}
M.~Malinowski and M.~Fritz.
\newblock A multi-world approach to question answering about real-world scenes
  based on uncertain input.
\newblock In {\em NIPS}, pages 1682--1690, 2014.

\bibitem{malinowski2014pooling}
M.~Malinowski and M.~Fritz.
\newblock A pooling approach to modelling spatial relations for image retrieval
  and annotation.
\newblock {\em arXiv preprint arXiv:1411.5190}, 2014.

\bibitem{mao2014deep}
J.~Mao, W.~Xu, Y.~Yang, J.~Wang, Z.~Huang, and A.~Yuille.
\newblock Deep captioning with multimodal recurrent neural networks (m-rnn).
\newblock In {\em ICLR}, 2015.

\bibitem{mao2014explain}
J.~Mao, W.~Xu, Y.~Yang, J.~Wang, and A.~L. Yuille.
\newblock Explain images with multimodal recurrent neural networks.
\newblock In {\em NIPS Deep Learning Workshop}, 2014.

\bibitem{mikolov2010recurrent}
T.~Mikolov, M.~Karafi{\'a}t, L.~Burget, J.~Cernock{\`y}, and S.~Khudanpur.
\newblock Recurrent neural network based language model.
\newblock In {\em INTERSPEECH}, pages 1045--1048, 2010.

\bibitem{mikolov2013distributed}
T.~Mikolov, I.~Sutskever, K.~Chen, G.~S. Corrado, and J.~Dean.
\newblock Distributed representations of words and phrases and their
  compositionality.
\newblock In {\em NIPS}, pages 3111--3119, 2013.

\bibitem{mitchell2012midge}
M.~Mitchell, X.~Han, J.~Dodge, A.~Mensch, A.~Goyal, A.~Berg, K.~Yamaguchi,
  T.~Berg, K.~Stratos, and H.~Daum{\'e}~III.
\newblock Midge: Generating image descriptions from computer vision detections.
\newblock In {\em EACL}, 2012.

\bibitem{ouyang2014deepid}
W.~Ouyang, P.~Luo, X.~Zeng, S.~Qiu, Y.~Tian, H.~Li, S.~Yang, Z.~Wang, Y.~Xiong,
  C.~Qian, et~al.
\newblock Deepid-net: multi-stage and deformable deep convolutional neural
  networks for object detection.
\newblock {\em arXiv preprint arXiv:1409.3505}, 2014.

\bibitem{papineni2002bleu}
K.~Papineni, S.~Roukos, T.~Ward, and W.-J. Zhu.
\newblock Bleu: a method for automatic evaluation of machine translation.
\newblock In {\em ACL}, pages 311--318, 2002.

\bibitem{ILSVRCarxiv14}
O.~Russakovsky, J.~Deng, H.~Su, J.~Krause, S.~Satheesh, S.~Ma, Z.~Huang,
  A.~Karpathy, A.~Khosla, M.~Bernstein, A.~C. Berg, and L.~Fei-Fei.
\newblock {ImageNet Large Scale Visual Recognition Challenge}, 2014.

\bibitem{salakhutdinov2010one}
R.~Salakhutdinov, J.~Tenenbaum, and A.~Torralba.
\newblock One-shot learning with a hierarchical nonparametric bayesian model.
\newblock 2010.

\bibitem{sharmanska2012augmented}
V.~Sharmanska, N.~Quadrianto, and C.~H. Lampert.
\newblock Augmented attribute representations.
\newblock In {\em ECCV}, pages 242--255. 2012.

\bibitem{simonyan2014very}
K.~Simonyan and A.~Zisserman.
\newblock Very deep convolutional networks for large-scale image recognition.
\newblock {\em arXiv preprint arXiv:1409.1556}, 2014.

\bibitem{socher2013zero}
R.~Socher, M.~Ganjoo, C.~D. Manning, and A.~Ng.
\newblock Zero-shot learning through cross-modal transfer.
\newblock In {\em NIPS}, pages 935--943, 2013.

\bibitem{socher2014grounded}
R.~Socher, Q.~Le, C.~Manning, and A.~Ng.
\newblock Grounded compositional semantics for finding and describing images
  with sentences.
\newblock In {\em TACL}, 2014.

\bibitem{sutskever2014sequence}
I.~Sutskever, O.~Vinyals, and Q.~V. Le.
\newblock Sequence to sequence learning with neural networks.
\newblock In {\em NIPS}, pages 3104--3112, 2014.

\bibitem{swingley2010fast}
D.~Swingley.
\newblock Fast mapping and slow mapping in children's word learning.
\newblock {\em Language learning and Development}, 6(3):179--183, 2010.

\bibitem{tommasi2014learning}
T.~Tommasi, F.~Orabona, and B.~Caputo.
\newblock Learning categories from few examples with multi model knowledge
  transfer.
\newblock {\em TPAMI}, 36(5):928--941, 2014.

\bibitem{vedantam2014cider}
R.~Vedantam, C.~L. Zitnick, and D.~Parikh.
\newblock Cider: Consensus-based image description evaluation.
\newblock {\em arXiv preprint arXiv:1411.5726}, 2014.

\bibitem{vinyals2014show}
O.~Vinyals, A.~Toshev, S.~Bengio, and D.~Erhan.
\newblock Show and tell: A neural image caption generator.
\newblock {\em arXiv preprint arXiv:1411.4555}, 2014.

\bibitem{weston2010large}
J.~Weston, S.~Bengio, and N.~Usunier.
\newblock Large scale image annotation: learning to rank with joint word-image
  embeddings.
\newblock {\em Machine learning}, 81(1):21--35, 2010.

\bibitem{xu2015show}
K.~Xu, J.~Ba, R.~Kiros, C.~A. Cho, Kyunghyun, R.~Salakhutdinov, R.~Zemel, and
  Y.~Bengio.
\newblock Show, attend and tell: Neural image caption generation with visual
  attention.
\newblock {\em arXiv preprint arXiv:1502.03044}, 2015.

\bibitem{zeiler2012adadelta}
M.~D. Zeiler.
\newblock Adadelta: an adaptive learning rate method.
\newblock {\em arXiv preprint arXiv:1212.5701}, 2012.

\bibitem{zhu2014learning}
J.~Zhu, J.~Mao, and A.~L. Yuille.
\newblock Learning from weakly supervised data by the expectation loss svm
  (e-svm) algorithm.
\newblock In {\em NIPS}, pages 1125--1133, 2014.

\end{thebibliography}
}

\end{document}